%% file: main.tex
\documentclass[preprint,5p,times,twocolum]{elsarticle}

\usepackage{amssymb}
\usepackage{amsmath}
\usepackage{amsthm}
\usepackage{microtype}

\usepackage{hyperref}

\journal{Neurocomputing}

\usepackage{algorithm}
\usepackage{algorithmicx}
\usepackage{algpseudocode}
\usepackage{graphicx}
\usepackage{booktabs}
\input{math_command}

\begin{document}

\begin{frontmatter}



\title{Enhancing Performance of 3D Point Completion Network using Consistency Loss}


% \author[kaist_master_of_robotics_program]{Christofel Rio Goenawan \corref{same_contribution_authors}}
% \ead{christofel.goenawan@kaist.ac.kr}
\author[maxplanc]{Kevin Tirta Wijaya \corref{same_contribution_authors}}
\ead{kevintirta.w@gmail.com}

\author[kaist_master_of_robotics_program]{Christofel Rio Goenawan \corref{same_contribution_authors}}
\ead{christofel.goenawan@kaist.ac.kr}
\author[kaist_cho_cun_sik_graduate_school_of_mobillity]{Seung-Hyun Kong\corref{cor}}
\ead{skong@kaist.ac.kr}

\affiliation[maxplanc]{organization={Computer Graphics Department, Max Planck Institute for Informatics},
            city={Saarbrücken},
            postcode={66123},
            country={Germany}}
\affiliation[kaist_master_of_robotics_program]{organization={Robotics Master Program, KAIST},%Department and Organization
            city={Daejeon},
            postcode={34051}, 
            country={Republic of Korea}}
\affiliation[kaist_cho_cun_sik_graduate_school_of_mobillity]{organization={CCS Graduate School of Mobility, KAIST},%Department and Organization
            city={Daejeon},
            postcode={34051}, 
            country={Republic of Korea}}

%\cortext[same_contribution_authors]{Authors equally contributed}
%\cortext[same_contribution_authors]{Both authors contributed equally in the project.}
\cortext[cor]{Corresponding author.}

%% Abstract
\begin{abstract}
Point cloud completion networks are conventionally trained to minimize the disparities between the completed point cloud and the ground-truth counterpart. However, an incomplete object-level point cloud can have multiple valid completion solutions when it is examined in isolation. This one-to-many mapping issue can cause contradictory supervision signals to the network because the loss function may produce different values for identical input-output pairs of the network. In many cases, this issue could adversely affect the network optimization process.
In this work, we propose to enhance the conventional learning objective using a novel completion consistency loss to mitigate the one-to-many mapping problem. Specifically, the proposed consistency loss ensure that a point cloud completion network generates a coherent completion solution for incomplete objects originating from the same source point cloud.
Experimental results across multiple well-established datasets and benchmarks demonstrated the proposed completion consistency loss have excellent capability to enhance the completion performance of various existing networks without any modification to the design of the networks. The proposed consistency loss enhances the performance of the point completion network without affecting the inference speed, thereby increasing the accuracy of point cloud completion. Notably, a state-of-the-art point completion network trained with the proposed consistency loss can achieve state-of-the-art accuracy on the challenging new MVP dataset. The code and result of experiment various point completion models using proposed consistency loss will be available at: https://github.com/kaist-avelab/ConsistencyLoss . 
\end{abstract}

%Graphical abstract
\begin{graphicalabstract}
\includegraphics{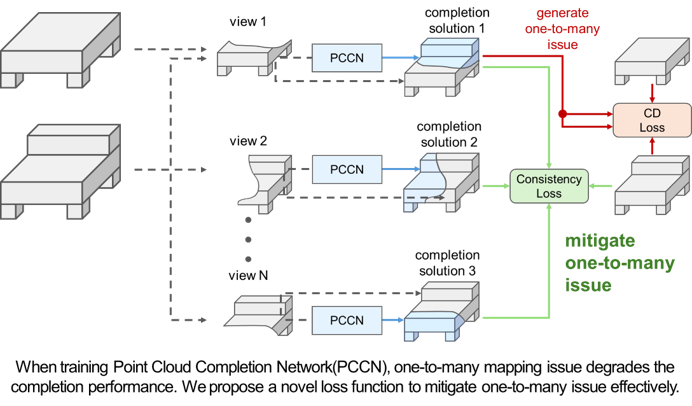}
\end{graphicalabstract}

{
\onecolumn
\begin{highlights}
\item Introduction of Completion Consistency loss: A novel loss function for Point Cloud Completion Networks (PCCNs) to address the one-to-many problem 
\item Compatibility with Existing Networks: The proposed consistency loss can be seamlessly integrated into existing PCCNs without any modification of their design.
\item Improved Network Performance and Efficiency: The proposed consistency loss significantly improved the performance of PCCNs, enabling simper network to achieve comparable to more complex networks.
\item Point Completion Network trained using proposed consistency loss on challenging MVP dataset achieve state-of-the-art performance for MVP dataset since MVP dataset contains high- quality multi-view point clouds from same ground truths.
%\item Point Completion Network trained using proposed consistency loss can predict missing point cloud using incomplete point clouds with generated noise more accurate than point completion network trained without proposed consistency loss on input incomplete point clouds with generated noise
\item The Point Completion Network trained with the proposed consistency loss more accurately predicts missing points in incomplete point clouds with generated noise compared to a network trained without the consistency loss.
\item Enhanced Generalization Capability: The proposed consistency loss also enhanced the network's ability to generalize to previously unseen objects.
\end{highlights}
}

%% Keywords

\end{frontmatter}

%%\linenumbers


\section{Introduction}
\label{sec:introduction}
%% Labels are used to cross-reference an item using \ref command.
Point cloud completion is a 3D reconstruction task of  occluded or incomplete point clouds. Point cloud completion is very important in 3D computer vision for such us 3D object detection \cite{qi2017pointnet} \cite{qi2017pointnet++} for robots especially drivable area detection \cite{Goenawan_Christofel_See_the_Unseen_Grid_Wise_Drivable_Area_Detection} and 3D object tracking \cite{mao20233dobjectdetectionautonomous} \cite{Qian_2022_Survey_of_3D_Object_Detection_for_Autonomous_Driving} \cite{goenawan2024astmautonomoussmarttraffic} for autonomous vehicles.  In recent years, numerous studies \cite{yang2018foldingnet, tchapmi2019topnet, huang2020pf_pfnet, wen2021pmp, chen2023anchorformer} have been conducted to leverage deep neural networks to complete occluded object-level point clouds\footnote{we use the terms "object" and "point cloud" interchangeably to refer to object-level point clouds}.
These point cloud completion networks (PCCNs) are often designed to take locally-incomplete point clouds as input and generate complete point clouds as output.

% \begin{figure}[ht]
%     \centering
%     \includegraphics[width=0.45\textwidth]{figure_1.pdf}
%     \caption{Contradictory supervision signals could appear when an incomplete point cloud have multiple possible completion solutions, and could lead the network to fall into suboptimal solution regions. Point clouds are represented with solid lines in the figure for clarity.}
%     \label{fig:contradictory_supervision_signal}
% \end{figure}

Improvements of the completion performance of recent PCCNs can primarily be attributed to innovations in network architectures \cite{yuan2018pcn, yu2021pointr, zhang2022attention_axform}, point generation strategies \cite{xiang2021snowflakenet, tang2022lakenet, wen2022pmpnet++}, and representations \cite{zhou2022seedformer}.
In contrast, the training strategy employed by existing PCCNs has remained relatively unchanged, that is, to minimize the dissimilarities between the predicted complete point clouds and the ground truths \cite{fei2022comprehensive_review_pccn} \cite{Enhancing_Performance_of_3D_Point_Completion_Network_using_Consistency_LOss}, often measured using the computationally efficient Chamfer Distance (CD) metric \cite{fan2017point_chamfer_distance_metric}.
Unfortunately, the straightforwardness of such a training strategy is not without a potential drawback: an incomplete point cloud, when inspected independently without additional information, could have multiple valid solutions according to the CD metric.

\begin{figure}[ht]
    \centering
    \includegraphics[width=0.45\textwidth]{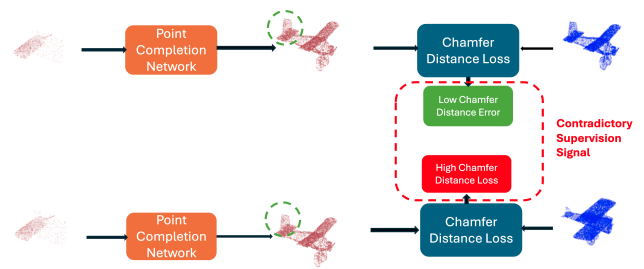}
    \caption{Contradictory supervision signals could appear when an incomplete point cloud have multiple possible completion solutions, and could lead the network to fall into suboptimal solution regions. Point clouds are represented with solid lines in the figure for clarity.}
    \label{fig:contradictory_supervision_signal}
\end{figure}

To illustrate, consider a simple scenario in which an incomplete point cloud has a partial cuboid shape, shown in Figure \ref{fig:contradictory_supervision_signal}.
This incomplete point cloud can be obtained from various objects such as a table, a bed, or other type of objects.
Such scenarios can lead to contradictory supervision signals during the training process, in which the loss function could yield various values for the same input-output pairs.
As a result, at the end of the training process, the network might produce middle-ground-solutions for both inputs that are suboptimal in terms of completion quality.

In this paper, we propose a novel completion consistency loss that can be easily integrated into the commonly-used training strategy, without any changes to the design of the networks.
The core idea of the completion consistency loss is to examine multiple incomplete views of a source object at the same time instead of inspecting them independently.
That is, at each forward-backward pass, we sample a set of incomplete point clouds originating from the same object, and take a gradient descent step with considerations to the fact that the completion solutions for each element in this set should be identical.
This is in contrast to the conventional training strategy, in which only one incomplete point cloud is considered for each source object at each forward-backward pass.

To demonstrate the effectiveness of the completion consistency loss, we evaluate three existing PCCNs,  PCN \cite{yuan2018pcn}, AxFormNet \cite{zhang2022attention_axform}, and AdaPoinTr \cite{yu2023adapointr}, on well-established benchmarks \cite{yuan2018pcn, yu2021pointr}, without any modifications to the original network architectures.
In all three networks, the completion performance is improved when the completion consistency loss is used during the training.
Furthermore, we observe that relatively fast but simple PCCNs (PCN and AxFormNet) that are trained with the consistency loss can match the completion accuracy of more complex but slower PCCNs.
In addition, experimental results demonstrated that the consistency loss can improve the capability of the networks to generalize to previously-unseen shape categories.
Therefore, the consistency loss could pave the way for accurate, fast, and robust PCCNs, especially for completing a set of point clouds with diverse shapes.

The remainder of this paper is as follows. Section \ref{sec:related_work} introduces background of this paper including related works. Section \ref{sec:sec3} explains consistency loss that is the main contribution of this paper. Section \ref{sec:sec4} discusses the experimental results. Section \ref{sec:sec5} concludes the paper with a summary.

\section{Background}
\subsection{Related Work}
\label{sec:related_work}
Traditional approaches \cite{wu20153d, dai2017shape, girdhar2016learning, han2017high} for 3D shape completion task often use voxels as the data representation.
However, the memory requirement for voxel-based operations grows cubically with respect to the spatial resolution.
In contrast, point cloud representation is capable of preserving 3D structure details with low memory requirement, and has become widely-used in many deep learning applications owing to the pioneering works of \cite{qi2017pointnet} and \cite{qi2017pointnet++}. 

PCN \cite{yuan2018pcn} is one of the first deep learning-based neural networks for point cloud completion.
It utilizes an encoder-decoder-folding scheme to learn features from the partial point cloud, and predicts the final reconstructed points with FoldingNet \cite{yang2018foldingnet}. 
Since then, numerous network architectures for point cloud completion have been proposed.
For example, TopNet \cite{tchapmi2019topnet} utilized softly-constrained decoder that is capable of generating point clouds based on a hierarchical rooted tree structure, and GRNet \cite{xie2020grnet} leveraged gridding operations to enable point cloud to 3D grids transformation without loss of structural information. 

Recently, attention-based architectures have grown in popularity as the go-to architecture for PCCN.
For example, PoinTr \cite{yu2021pointr} use a geometry-aware transformer architecture to estimate coarse point predictions before performing refinement via FoldingNet \cite{yang2018foldingnet}, while Seedformer \cite{zhou2022seedformer} introduces Patch Seeds as a new shape representation which contains seed coordinates and features of a small region in the point cloud. %PointAttN \cite{Wang_Cui_Guo_Li_Liu_Shen_2024_PointAttn} proposed incomplete point cloud prediction using fast- computation attention for point clouds. While AdapointTr \cite{yu2023adapointr} try to predict unseen point clouds based on the adaptive geometry module and SVDFormer \cite{zhu2023svdformercomplementingpointcloud} detects unseen point clouds based on the partial point clouds and multi view image projection of the partial point clouds. While Wu et al \cite{wu2021density_dcd_loss} try to proposed density-aware Chamfer Distance loss for predicting unseen point clouds well on area without dense known point clouds.

Zhiang et al. proposed detecting unseen point clouds from partial point clouds using a multi-stage points prediction approach based on point density \cite{zhang2020preservedpointcloudcompletion}. The MSCPN method attempts to detect unseen point clouds in multiple stages, utilizing the geometric features of the partial point clouds \cite{ZHANG2020101925_Multi_Stage_Point_Completion_Network_with_Critical_Set_Supervision}. Wang et al. proposed predicting unseen point clouds by predicting the geometric edges of the unseen point clouds from the geometric features of the partial point clouds \cite{wang2021voxelbasednetworkshapecompletion}. Zhang et al. suggested predicting unseen point clouds using the geometric shape of the unseen point clouds with skeleton-detailed transformer models \cite{9804851_Point_Cloud_Completion_via_Skeleton_Detail_Transformer}. PointAttN \cite{Wang_Cui_Guo_Li_Liu_Shen_2024_PointAttn} introduced a method for predicting incomplete point clouds using fast-computation attention mechanisms for point clouds. AdapointTr \cite{yu2023adapointr} aims to predict unseen point clouds by utilizing an adaptive geometry module, while SVDFormer \cite{zhu2023svdformercomplementingpointcloud} predicts unseen point clouds based on partial point clouds and multi-view image projections of the partial point clouds. SOE-Net \cite{29} introduces a self-attention and orientation encoding network for point cloud-based place recognition. It employs a PointOE module to capture local information from eight orientations and a self-attention unit to encode long-range feature dependencies, achieving state-of-the-art performance in place recognition tasks.

FSC proposed a novel method for predicting unseen point clouds from very small partial point clouds using a two-stage refinement process for global and local geometric feature prediction. Wu et al. \cite{wu2021density_dcd_loss} introduced a novel density-aware Chamfer Distance loss to improve the prediction of unseen point clouds, particularly in regions lacking dense known point clouds. Recent advancements include CASSPR \cite{Xia_2023_ICCV_CASSPR_Cross_Attention_Single_Scan_Place_Recognition}, which improves LiDAR-based place recognition by combining point-based and voxel-based methods with a hierarchical cross-attention mechanism. Another approach, ASFM-Net \cite{10.1145/3474085.3475348_ASFM_Net_Assymetrical_Siamese_Feature_Matching_Network_for_Point_Completion}, introduces an asymmetrical Siamese feature matching network for point completion, utilizing iterative refinement to generate complete shapes from partial inputs.

Recently, some researchers have attempted to predict unseen point clouds from partial point clouds using generative models, particularly diffusion-based models \cite{ho2020denoisingdiffusionprobabilisticmodels}. Lyu et al. proposed generating unseen point clouds from partial point clouds using a conditional diffusion model with the Point Diffusion Paradigm \cite{lyu2022conditionalpointdiffusionrefinementparadigm}. Zheng et al. suggested using a diffusion model to train point cloud completion networks, leveraging pretrained models from various point cloud completion networks trained on different datasets \cite{zheng2023pointcloudpretrainingdiffusion}. Karsten et al. introduced the use of a diffusion model with a text-to-image pretrained model to predict unseen point clouds from semantic patterns in partial point clouds \cite{NEURIPS2023_284afdc2_Point_Cloud_Completion_with_Pretrained_Text_to_Image_Diffusion_Models}. Yuhan et al. proposed 3DQD, which generates unseen point clouds using a conditional diffusion model based on geometric features of partial point clouds \cite{Li_2023_CVPR_Generalized_Deep_3D_Shape_Prior_via_Part_Discretized_Difussion_Process}. Romanelis et al. introduced a novel fusion of point-based and voxel-based feature extraction to generate unseen point clouds using conditional diffusion models \cite{romanelis2024efficientscalablepointcloud}.
While diffusion models can learn the underlying patterns of point clouds from partial point clouds, point cloud completion using diffusion models often requires significant time and computational resources for both training and prediction. In contrast, our proposed consistency loss enables point completion networks to learn features of point clouds from the same ground truth more effectively. As a result, point completion networks trained with our proposed consistency loss can predict unseen point clouds from partial point clouds more accurately, without increasing the prediction time.

% Shi et al proposed accurate 3D object detection for autonomous vehicle by completing 3D point clouds using point completion network \cite{Shi_2022_ACCV}. Yufei et al proposed novel 3D point cloud completion for outdoor RGB-D data for accurate 3D object detection by combining LiDAR point clouds and RGB images \cite{9428349_LPCCNet_RGB_Guided_Local_Point_Cloud_Completion}. Shan et al proposed 3D object detection for autonomous vehicle using very few labelled 3D object using point completion network on LiDAR point clouds \cite{shan2023scpscenecompletionpretraining}. While Liang et al proposed accurate 3D object detection using point cloud completion to predict 3D object in remote area \cite{LIANG2024104049_Boosting_3D_point_based_object_detection}. Koo et al proposed novel 3D object detection using surface point cloud completion for accurate 3D object detection \cite{koo2023pgrcnnsemanticsurfacepoint}. Tang et al proposed generated point cloud corner of 3D objects for accurate 3D object detection \cite{TANG2024112117_Boundary_Guided_3D_Object_Detection_for_Point_Clouds}. 

Shi et al. proposed an approach for accurate 3D object detection in autonomous vehicles by completing 3D point clouds using a point completion network \cite{Shi_2022_ACCV}. Yufei et al. introduced a novel method for 3D point cloud completion in outdoor RGB-D data, enhancing 3D object detection accuracy by combining LiDAR point clouds and RGB images \cite{9428349_LPCCNet_RGB_Guided_Local_Point_Cloud_Completion}. Shan et al. developed a 3D object detection method for autonomous vehicles using point completion networks on LiDAR point clouds, requiring only a few labeled 3D objects \cite{shan2023scpscenecompletionpretraining}. Liang et al. proposed a model that uses point cloud completion to improve 3D object detection in remote areas \cite{LIANG2024104049_Boosting_3D_point_based_object_detection}. Koo et al. introduced a novel approach using surface point cloud completion for precise 3D object detection \cite{koo2023pgrcnnsemanticsurfacepoint}. Lastly, Tang et al. proposed a technique to generate point cloud corners of 3D objects to improve detection accuracy \cite{TANG2024112117_Boundary_Guided_3D_Object_Detection_for_Point_Clouds}.

\subsection{Preliminary Findings}

\textbf{Optimal training strategy can improve completion performance.}
The works discussed in Subsection \ref{sec:related_work} mainly focus on architectural innovations to improve the state-of-the-art point cloud completion performance.
On the other hand, several works \cite{liu2022convnet_convnext, qian2022pointnext, steiner2022how_howtotrainyourvit} have highlighted that a well-designed training strategy can improve the performance of a neural network.
As such, we posit that developing a good training strategy could yield similar advantages for the completion performance of PCCNs.

A training strategy covers a wide array of aspects including the choice of optimizer, learning rate schedule, regularization techniques, data augmentations, auxiliary tasks, and more.
To emphasize the significance of a well-designed training strategy, we train a PCN \cite{yuan2018pcn} model using the AdamW \cite{loshchilov2017decoupled_adamw} optimizer for 250 epochs, with a cosine annealing \cite{loshchilov2016sgdr_cosineannealing} scheduler.
We set the maximum and minimum learning rates to $10^{-4}$ and $5 \cdot 10^{-5}$, respectively, and keep the network architecture and other hyperparameters identical with those used by \cite{yu2021pointr}.

As shown in Table \ref{tab:pcn_improved_training_strategy}, the PCN model trained with this improved strategy achieved a CD$_{l2}$ score of $2.37\cdot 10^{-3}$, a substantial improvement over the previously reported performance of $4.08 \cdot 10^{-3}$, and closer to the completion performance of more recent transformer-based models such as PoinTr \cite{yu2021pointr}.
This result clearly demonstrates the positive impacts of a good training strategy to the completion performance of a PCCN.

\begin{table}[t]
  \centering
  \caption{Completion performance on ShapeNet55-hard where 75\% of the original points are missing. $^1$As reported in the ShapeNet55 benchmark \cite{yu2021pointr}.\\}
  \begin{tabular}{lc}
      \toprule
      \multicolumn{1}{c}{Model} & CD$_{l2} \times 10^3$  $\downarrow$\\
      \cmidrule{1-2}
      PCN$^1$ & 4.08 \\
      + Improved Training & 2.37\\
%      \hspace{1.0em} Training Strategy & \\
      PoinTr$^1$ & 1.79 \\
      \bottomrule
  \end{tabular}
  \label{tab:pcn_improved_training_strategy}
\end{table}

\textbf{Learning to predict only the missing points can improve completion performance.}
\label{sec:common_training_objective}
%Another aspect of training strategy for PCCNs is the formulation of the point cloud completion problem.
In the literature, there are at least two major problem formulation for deep learning-based point cloud completion.
Let ${\mathbb{P}}^{\text{com}}$ be a set of points $p^{\text{com}}_i \in {\mathbb{P}}^3$ sampled from an object $O$ and $\Phi$ be a neural network.
We can obtain two disjoint sets from ${\mathbb{P}}^{\text{com}}$: the set of missing points ${\mathbb{P}}^{\text{mis}}$ and the set of incomplete points ${\mathbb{P}}^{\text{inc}}$, where ${\mathbb{P}}^{\text{com}} = {\mathbb{P}}^{\text{mis}} \cup {\mathbb{P}}^{\text{inc}}$ and ${\mathbb{P}}^{\text{mis}} \cap {\mathbb{P}}^{\text{inc}} = \emptyset$.

In the first approach \cite{yuan2018pcn, zhang2022attention_axform}, the goal is to estimate the entire complete point cloud given an incomplete point cloud, $\Phi({\mathbb{P}}^{\text{inc}}) = \hat{{\mathbb{P}}}^{\text{com}}$ and minimize the completion error as measured by the Chamfer Distance, $\text{CD}(\hat{{\mathbb{P}}}^{\text{com}}, {\mathbb{P}}^{\text{com}})$.
In the second approach \cite{yu2021pointr}, the goal is to estimate only the missing point cloud given an incomplete point cloud, $\Phi({\mathbb{P}}^{\text{inc}}) = \hat{{\mathbb{P}}}^{\text{mis}}$ and minimize $\text{CD}(\hat{{\mathbb{P}}}^{\text{mis}}, {\mathbb{P}}^{\text{mis}})$.
The estimated complete point cloud of the second approach is then the union of the predicted missing points and the input incomplete points, 
$\hat{{\mathbb{P}}}^{\text{com}} = \hat{{\mathbb{P}}}^{\text{mis}} \cup {\mathbb{P}}^{\text{inc}}$.

\begin{table}[t]
  \centering
  \caption{Completion performance on ShapeNet55-hard where 75\% of the original points are missing. We use AxForm \cite{zhang2022attention_axform} as $\Phi$.\\}
  \begin{tabular}{lc}
      \toprule
      \multicolumn{1}{c}{Model} & CD$_{l2} \times 10^3$  $\downarrow$\\
      \cmidrule{1-2}
      $\Phi(\mathbb{P}^{\text{inc}}) = \hat{\mathbb{P}}^{\text{mis}}$ & 1.62 \\
      $\Phi(\mathbb{P}^{\text{inc}}) = \hat{\mathbb{P}}^{\text{com}}$ & 1.80 \\
      \bottomrule
  \end{tabular}
  \label{tab:axform_missing_prediction}
\end{table}

To compare the completion performance between the two approaches, we train two AxForm networks \cite{zhang2022attention_axform}, one for each approach.
% From our experiments, the second approach of predicting only the missing points yields better completion performance (CD$_{l2} = 1.62 \times 10^{-3}$) than the first approach of predicting the complete points (CD$_{l2} = 1.80 \times 10^{-3}$.)
As shown in \ref{tab:axform_missing_prediction}, the second approach (predicting only the missing points) yields better completion performance than the first approach (predicting complete points).
Therefore, the experiments in the following sections are based on the second approach, for which the objective can be considered as a reconstruction loss,
\begin{equation}
\mathcal{L}^{\text{rec}}_k = \text{CD}(\hat{{\mathbb{P}}}^{\text{mis}}_k, {\mathbb{P}}^{\text{mis}}_k),   
\label{eq:rec_loss}
\end{equation}
where CD is defined as,
\begin{equation}
\begin{aligned}
&\text{CD}({\mathbb{A}}, {\mathbb{B}})\\ &= \frac{1}{|{\mathbb{A}}|}\sum_{{\bm{a}} \in {\mathbb{A}}}\min_{{\bm{b}} \in {\mathbb{B}}} ||{\bm{a}} - {\bm{b}}||^2_2 + \frac{1}{|{\mathbb{B}}|}\sum_{{\bm{b}} \in {\mathbb{B}}}\min_{{\bm{a}} \in {\mathbb{A}}} ||{\bm{b}} - {\bm{a}}||^2_2.    
\end{aligned}
\label{eq:cd_metric}
\end{equation}

\textbf{One-to-many mapping issue can worsen the completion performance.}
\label{sec:pitfall}
To investigate the potential impact of the one-to-many mapping issue on the completion performance of PCCNs, we conduct experiments on toy datasets that are derived from the Shapenet55 dataset.
First, we construct two types of toy datasets, ${\mathbb{D}}^{A} = \bigcup_{i=1}^{5} {\mathbb{D}}_i^{A}$ and ${\mathbb{D}}^{B} = \bigcup_{i=1}^{5} {\mathbb{D}}_i^{B}$, where ${\mathbb{D}}_i^A$ and ${\mathbb{D}}_i^B$ each consists of 5,000 samples from ShapeNet55.
The samples in ${\mathbb{D}}^{A}_i$ is selected in a way such that, on average, the CD-score between ${\mathbb{P}}^{\text{inc}}_j \in {\mathbb{D}}^{A}_i$ and ${\mathbb{P}}^{\text{inc}}_k \in {\mathbb{D}}^{A}_i$, $j \neq k$, is relatively low, but the CD-score between ${\mathbb{P}}^{\text{mis}}_j \in {\mathbb{D}}^{A}_i$ and ${\mathbb{P}}^{\text{mis}}_k \in {\mathbb{D}}^{A}_i$, $j \neq k$, is relatively high.
% Further details regarding the steps to generate ${\mathbb{D}}^{A}$ can be found in Appendix \ref{sec:generating_toy_datasets}.
Meanwhile, samples in ${\mathbb{D}}^{B}_i$ are randomly selected from Shapenet55 with uniform probabilities and therefore is statistically similar to the full ShapeNet55 dataset.

\begin{table}[ht]
      \centering
      \caption{Completion performance on Toy Datasets based on ShapeNet55-hard.\\}
      \begin{tabular}{lc}
          \toprule
          \multicolumn{1}{c}{Model} & CD$_{l2} \times 10^3$  $\downarrow$\\
          \cmidrule{1-2}
          AxForm on ${\mathbb{D}}^{A}$ & 2.81 ± 0.15 \\
          AxForm on ${\mathbb{D}}^{B}$ & 2.44 ± 0.10 \\
          \bottomrule
      \end{tabular}
  \label{tab:toydataset}
\end{table}

We use 80\% of the samples in each dataset for training, and hold the remaining 20\% for evaluation.
In total, we train 10 AxForm networks \cite{zhang2022attention_axform} on ${\mathbb{D}}^A$ and ${\mathbb{D}}^B$, and report the average and standard deviation of the CD-scores.
As shown in Table \ref{tab:toydataset}, the CD-score of networks trained and evaluated on ${\mathbb{D}}^B$ is lower (better) than the CD-score of networks trained and evaluated on ${\mathbb{D}}^A$.
These results indicate that the one-to-many mapping issue negatively affects the completion performance of the PCCNs.

\section{Consistency Loss}
\label{sec:sec3}

\begin{figure}[ht]
    \centering
    \includegraphics[width=0.45\textwidth]{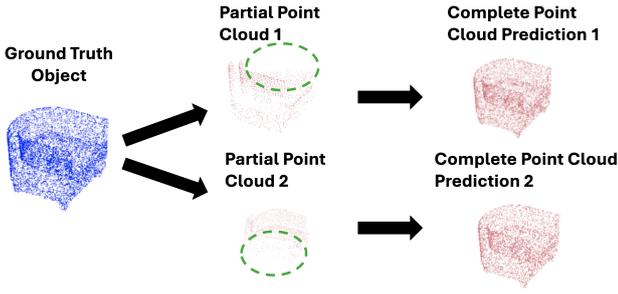}
    \caption{Two different incomplete point clouds that are obtained from one object should have the same solutions. Point completion network try to predict unseen point clouds on green dashed circle area. }
    \label{fig:one_object_same_solution}
\end{figure}

In this section, we introduce the completion consistency loss, which we refer to as the consistency loss for brevity from here onward, to mitigate the aforementioned issues.
The core idea of the consistency loss is to consider multiple incomplete point clouds originating from the same source object before taking a gradient descent step (Figure \ref{fig:one_object_same_solution}).
Recall that the contradictory supervision signals exist when there are multiple valid completion solutions for one incomplete point cloud that is observed in isolation.
Therefore, intuitively, adding more incomplete point clouds with the same completion solution at one observation can reduce the ambiguity and mitigate the negative effects of the issue.

% \subsection{Completion consistency Loss}
We propose two ways to implement the consistency loss: self-guided consistency and target-guided consistency.

\subsection{Self-guided Consistency}
In self-guided consistency loss, we leverage the fact that we can generate multiple incomplete point clouds from the same object, and utilize these samples in the consistency loss.
Given a complete point cloud ${\mathbb{P}}^{\text{com}}_k$ representing the object $k$, we can generate a set of $n$ different incomplete point clouds ${\mathbb{P}}^{\text{inc}}_k = \{{\mathbb{P}}^{\text{inc}}_{k, 1}, {\mathbb{P}}^{\text{inc}}_{k, 2}, ..., {\mathbb{P}}^{\text{inc}}_{k, n}\}$.
Since the source of all incomplete point clouds is the same, that is, ${\mathbb{P}}^{\text{com}}_k$, the completion solutions for all ${\mathbb{P}}^{\text{inc}}_{k, i}$ should also be the same.
Therefore, given $\Phi({\mathbb{P}}^{\text{inc}}_{k, i}) = \hat{\mathbb{P}}^{\text{mis}}_{k,i}$ and $\hat{\mathbb{P}}^{\text{com}}_{k, i} = \hat{\mathbb{P}}^{\text{mis}}_{k,i} \cup {\mathbb{P}}^{\text{inc}}_{k,i}$, we can guide the network to produce similar completion solutions for any incomplete point clouds originating from ${\mathbb{P}}^{\text{com}}_k$ through the self-guided consistency,

\begin{equation}
\mathcal{L}^{\text{c-sg}}_k = \frac{2}{n(n-1)} \sum_{i=1}^{n-1}\sum_{j = i+1}^{n} \text{CD}(\hat{{\mathbb{P}}}^{\text{com}}_{k,i}, \hat{{\mathbb{P}}}^{\text{com}}_{k,j}).
\end{equation}

\subsection{Target-guided Consistency}
For target-guided consistency, we utilize the original ground truth for the consistency loss.
As mentioned in Subsection \ref{sec:common_training_objective}, the commonly-used loss function is calculated as either $\text{CD}(\Phi({\mathbb{P}}^{\text{inc}}), {\mathbb{P}}^{\text{com}})$ or $\text{CD}(\Phi({\mathbb{P}}^{\text{inc}}), {\mathbb{P}}^{\text{mis}})$.
While it turned out that the $\text{CD}(\Phi({\mathbb{P}}^{\text{inc}}), {\mathbb{P}}^{\text{mis}})$ is advantageous to the completion performance of PCCNs, the formulation does not promote consistency between completions because the supervision is only performed on ${\mathbb{P}}^{\text{mis}}$ instead of ${\mathbb{P}}^{\text{com}}$.
In target-guided consistency, we propose to keep the approach of predicting only the missing points, but we calculate the loss values based on the full complete point clouds.
Specifically, given a complete point cloud ${\mathbb{P}}^{\text{com}}_k$, $\Phi({\mathbb{P}}^{\text{inc}}_{k, i}) = \hat{\mathbb{P}}^{\text{mis}}_{k,i}$ and $\hat{\mathbb{P}}^{\text{com}}_{k, i} = \hat{\mathbb{P}}^{\text{mis}}_{k,i} \cup {\mathbb{P}}^{\text{inc}}_{k,i}$, the target-guided consistency is defined as,
\begin{equation}
\mathcal{L}^{\text{c-tg}}_k = \frac{1}{n} \sum_{i=1}^{n} \text{CD}(\hat{\mathbb{P}}^{\text{com}}_{k, i}, {\mathbb{P}}^{\text{com}}_k).
\end{equation}

%Intuitively, the optimizer would take into consideration that objects originating from the same source should have the same completion solution when performing a single gradient decent step.

\subsection{Complete Loss Function}
The complete loss function for a complete point cloud ${\mathbb{P}}^{\text{com}}_k$ with $n$ samples of incomplete point clouds is the combination of conventional reconstruction loss, self-guided consistency loss, and target-guided consistency loss, with scaling factors $\alpha \text{ and } \beta$,

\begin{equation}
    \mathcal{L}^{\text{total}}_k = \alpha \mathcal{L}^{\text{c-sg}}_k + \beta \mathcal{L}^{\text{c-tg}}_k + \frac{1}{n}\sum_{i=1}^{n} \mathcal{L}^{\text{rec}}_{i,k},
    \label{eq:loss_function}
\end{equation}
where $\mathcal{L}_{i, k}^{\text{rec}}$ is the reconstruction loss (Equation \ref{eq:rec_loss}) for $\hat{\mathbb{P}}^{\text{mis}}_{k, i}$.

We can also further improve the performance of the Point Cloud completion using density- aware chamfer distance loss as \cite{wu2021density_dcd_loss} with weight of density-aware chamfer distance loss $\delta$ as below.

\begin{equation}
    \mathcal{L}^{\text{total, da}}_k = \alpha \mathcal{L}^{\text{c-sg}}_k + \beta \mathcal{L}^{\text{c-tg}}_k + \frac{1}{n}\sum_{i=1}^{n} \mathcal{L}^{\text{rec}}_{i,k}+ \delta \frac{1}{n} \mathcal{L}^{\text{da}}_{i,k},
    \label{eq:loss_function_with_density_aware_chamfer_distance}
\end{equation}
where density-aware chamfer distance loss can be defined from \cite{wu2021density_dcd_loss} as below.

\begin{equation}
\begin{aligned}
&\text{DA-CD}({\mathbb{A}}, {\mathbb{B}})\\ &= \frac{1}{|{\mathbb{A}}|}\sum_{{\bm{a}} \in {\mathbb{A}}}\min_{{\bm{b}} \in {\mathbb{B}}} (1- e^{-||{\bm{a}} - {\bm{b}}||_2}) + \frac{1}{|{\mathbb{B}}|}\sum_{{\bm{b}} \in {\mathbb{B}}}\min_{{\bm{a}} \in {\mathbb{A}}} (1- e^{-||{\bm{b}} - {\bm{a}}||_2})
\end{aligned}
\label{eq:cd_metric}
\end{equation}

Note that both consistency losses do not directly eliminate the one-to-many mapping issue, but they can provide the network with additional information such that the network can mitigate the issue.
For a simple example, consider two inputs ${\mathbb{P}}^{\text{inc}}_{a, 1}$ and ${\mathbb{P}}^{\text{inc}}_{b, 1}$, and the corresponding completion solutions ${\mathbb{P}}^{\text{com}}_{a}$ and ${\mathbb{P}}^{\text{com}}_{b}$.
Suppose that CD$({\mathbb{P}}^{\text{inc}}_{a, 1}, {\mathbb{P}}^{\text{inc}}_{b, 1}) \approx 0$, and CD$({\mathbb{P}}^{\text{com}}_{a}, {\mathbb{P}}^{\text{com}}_{b}) >> \text{CD}({\mathbb{P}}^{\text{inc}}_{a, 1}, {\mathbb{P}}^{\text{inc}}_{b, 1})$, that is, the inputs are similar but the ground truths are dissimilar.
Assuming that $\Phi({\mathbb{P}}^{\text{inc}}_{a, 1})$ is also similar to $\Phi({\mathbb{P}}^{\text{inc}}_{b, 1})$, then a contradictory supervision signal could arise when we only use $\mathcal{L}^{\text{rec}}$ as the loss function.
On the other hand, suppose that we supplement the loss function with the consistency loss with $n=3$ such that the inputs become $\{{\mathbb{P}}^{\text{inc}}_{a, 1}, {\mathbb{P}}^{\text{inc}}_{a, 2}, {\mathbb{P}}^{\text{inc}}_{a, 3}\}$ and $\{{\mathbb{P}}^{\text{inc}}_{b, 1}, {\mathbb{P}}^{\text{inc}}_{b,2}, {\mathbb{P}}^{\text{inc}}_{b,3} \}$ for each ground truth.
The effect of the contradictory supervision signal to the gradient descent step can then be suppressed by $\mathcal{L}^{\text{c-sg}}_k \text{ and } \mathcal{L}^{\text{c-tg}}_k$. 
We observe that the density-aware Chamfer Distance loss proposed by \cite{wu2021density_dcd_loss} allows the Chamfer Distance loss to predict unseen point clouds in areas without dense point clouds. In contrast, our proposed consistency loss enhances the point completion network's ability to predict unseen point clouds by detecting underlying point cloud patterns from the same ground truth objects. Therefore, our proposed self-guided and target-guided consistency losses for point completion can also leverage the density-aware Chamfer Distance loss to predict unseen point clouds from partial point clouds. Because proposed consistency loss increase performance of point cloud completion in training then point completion network can predict unseen point clouds with same prediction time with point cloud completion network trained without proposed consistency loss.
%We can see also that because density-aware Chamfer Distance loss by \cite{wu2021density_dcd_loss} provides option for Chamfer Distance loss to predict unseen point clouds on area without dense point clouds while our proposed consistency loss can make point completion network predicts unseen point clouds better by detecting underlying point cloud pattern from same ground truth objects then our proposed self-guided consistency loss for point completion and target- guided consistency loss for point completion can also used density-aware Chamfer Distance loss for predicting unseen point clouds from partial point clouds.
%However, suppose that we supplement the loss function with the consistency loss with $n=3$ such that the inputs become $\{{\mathbb{P}}^{\text{inc}}_{a, 1}, {\mathbb{P}}^{\text{inc}}_{a, 2}, {\mathbb{P}}^{\text{inc}}_{a, 3}\}$ and $\{{\mathbb{P}}^{\text{inc}}_{b, 1}, {\mathbb{P}}^{\text{inc}}_{b,2}, {\mathbb{P}}^{\text{inc}}_{b,3} \}$ for each ground truth.
%Assuming that CD(${\mathbb{P}}^{\text{inc}}_{a, i}, {\mathbb{P}}^{\text{inc}}_{b, j}$) is relatively high for all pair of $i \text{ and } j$ except for $i = j = 1$, then the effects of the contradictory supervision signals to the gradient descent step would be suppressed by the other terms.

\section{Experimental Results}
\label{sec:sec4}

In this section, we demonstrate the effectiveness of the consistency loss by comparing the completion performance of five existing PCCNs on four commonly-used datasets.
First, we explain the experimental setups that are needed to reproduce the results.
Then, we report and discuss the completion performance of three existing PCCNs trained with and without the consistency loss.
We also conduct additional experiments to check the effects of each component in the consistency loss.
All code will be available on the github page\footnote{https://github.com/kaist-avelab/ConsistencyLoss}.

\subsection{Experimental Setup}
\subsubsection{Datasets}
There are numerous object-level point clouds datasets, most of which are derived from the Shapenet dataset \cite{chang2015shapenet}, for example, PCN \cite{yuan2018pcn}, Completion3D \cite{tchapmi2019topnet}, and Shapenet55-34 \cite{yu2021pointr}. Last year Pan Liang et al also proposed novel point completion dataset MVP dataset \cite{pan2021variationalrelationalpointcompletion_mvp_point_completion_prediction_dataset} that consisted more than 100,000 incomplete point clouds with high- quality camera  with 26 generated partial point clouds from every ground truth point clouds. The dataset provide 26 generated partial point clouds from 16 different object classes hence will be very good to test our proposed consistency loss for point completion that will increase accuracy of point completion network to predict unseen point clouds from same ground truth object.
We choose to evaluate the consistency loss on the PCN, MVP dataset and Shapenet55-34 datasets, following \cite{yu2021pointr, zhou2022seedformer, yu2023adapointr}.

PCN consists of around 30K samples of point clouds, spanning over 8 categories: airplane, cabinet, car, chair, lamp, sofa, table, and vessel.
On the other hand, Shapenet55-34 consists of around 52K samples of point clouds from 55 categories, resulting in a considerably more diverse set of objects compared with PCN.
In Shapenet55, the dataset is split into 41,952 samples for training and 10,518 samples for evaluation, with samples from all 55 categories are present in both training and evaluation splits.
Meanwhile in Shapenet34, the dataset is split into 46,765 samples for training and 5,705 samples for evaluation, where the training split consists of samples from 34 categories, and the evaluation split consists of samples from all 55 categories.
Shapenet34 can be seen as an evaluation on out-of-distribution data since the 21 extra categories on the evaluation split are withheld during training.

\subsubsection{Implementation Details}
The consistency loss is designed to improve a PCCN without any modification to the architecture of the network.
Therefore, we used three existing PCCNs,  PCN \cite{yuan2018pcn}, AxFormNet \cite{zhang2022attention_axform}, PointAttN \cite{Wang_Cui_Guo_Li_Liu_Shen_2024_PointAttn}, SVDFormer \cite{zhu2023svdformercomplementingpointcloud} and AdaPoinTr \cite{yu2023adapointr} to evaluate the effectiveness of the consistency loss.
For fairness, we train two versions of all three PCCNs from scratch using publicly-available source codes and the same training strategy, e.g., identical problem formulation, optimizer, number of iterations, batch size, and learning rate schedule. The only difference between the two versions is that whether the consistency loss is incorporated into the loss function or not. For MVP dataset we only test proposed consistency loss for SVDFormer and AdaPointTr point completion network since only SVDFormer and AdaPointTr network that provides codes for training the model in MVP dataset to make sure our experiment training SVDFormer and AdaPointTr with consistency loss on MVP dataset same with original model .  We trained state-of-the-art models, including the Point Completion Network (PCN), SVDFormer, and AdaPointTr, on incomplete point clouds using random Gaussian noise with mean $0$ and standard deviation $0.01$, $\mathbb{N}(0, 0.01)$, applied to the $x$, $y$, and $z$ position axes. The training was performed on the benchmark Point Completion task dataset, the PCN dataset \cite{yuan2018pcn}. The goal of the training was to predict the missing points in the point cloud without noise.%We trained state-of-the-art Point Completion Network PCN, SVDFormer and AdaPointTr on incomplete point cloud using random noise on benchmark Point Completion task dataset PCN dataset \cite{yuan2018pcn} with random noise Gaussian Random Noise $=\mathbb{N}( 0, 0.01 )$ on point cloud position axis $x,y,z$ and try to predict missing point cloud without Random Noise.

All PCCNs are implemented with PyTorch \cite{paszke2019pytorch} and trained on RTX 3090 GPUs.
The batch sizes are set to 64, 64, and 16 for PCN, AxFormNet, and AdaPoinTr, respectively.
We set the number of epochs to 200, 400, and 600 for PCN, AxFormNet, and AdaPoinTr, respectively, utilize cosine annealing \cite{loshchilov2016sgdr_cosineannealing} for the learning rate schedule, and set $n = 3$ for the consistency loss.
We use Open3D \cite{Zhou2018_open3d} to visualize the point clouds.

\subsection{Main Results}

\subsubsection{Quantitative Results}

\begin{table*}[htb!]
    \centering
    \caption{Quantitative results on the PCN \cite{yuan2018pcn} and ShapeNet55 \cite{yu2021pointr} benchmarks. We report the L1-norm Chamfer Disctance (CD$_{l1}$) and L2-norm Chamfer Distance (CD$_{l2}$) for PCN and ShapeNet55, respectively. S, M, and H represent the simple, moderate, and hard setups, where the proportions of missing points are 25\%, 50\%, and 75\%, respectively. $^\dagger$ indicates that the models are trained from scratch based on source codes from \cite{yu2021pointr} , \cite{Wang_Cui_Guo_Li_Liu_Shen_2024_PointAttn}, \cite{zhu2023svdformercomplementingpointcloud} \cite{zhang2022attention_axform}, respectively.\\}
    \begin{tabular}{lccccc}
    \toprule
    & PCN & \multicolumn{4}{c}{ShapeNet55} \\
    & Avg. & S & M & H & Avg. \\
    & CD$_{l1}\times 10^3\downarrow$ & \multicolumn{4}{c}{CD$_{l2} \times 10^3$ $\downarrow$}\\
    \cmidrule(lr){1-6}
    FoldingNet \cite{yang2018foldingnet} & 14.31 & 2.67 & 2.66 & 4.05 & 3.12\\
    PCN$^\dagger$ \cite{yuan2018pcn} & 10.55 & 0.82 & 1.25 & 2.37 & 1.48\\%9.64\\
    \hspace{0.7em} + \textbf{Consistency Loss} & \textbf{10.52} & \textbf{0.54} & \textbf{0.93} & \textbf{1.74} & \textbf{1.07}\\
    TopNet \cite{tchapmi2019topnet} & 12.15 & 2.26 & 2.16 & 4.30 & 2.91\\
    GRNet \cite{xie2020grnet} & 8.83 & 1.35 & 1.71 & 2.85 & 1.97\\
    SnowflakeNet \cite{wen2021pmp} & 7.21 & 0.70 & 1.06 & 1.96 & 1.24\\
    PoinTr \cite{yu2021pointr} & 8.38 & 0.58 & 0.88 & 1.79 & 1.09\\
    AXFormNet$^\dagger$ \cite{zhang2022attention_axform} & & 0.72 & 1.06 & 1.98 & 1.22  \\
    \hspace{0.7em} + \textbf{Consistency Loss} & & \textbf{0.45} & \textbf{0.79} & \textbf{1.51} & \textbf{0.91} \\
    SeedFormer \cite{zhou2022seedformer} & 6.74 & 0.50 & 0.77 & 1.49 & 0.92\\
    PointAttn $^\dagger$\cite{Wang_Cui_Guo_Li_Liu_Shen_2024_PointAttn}& 6.84 & 0.47 & 0.66 & 1.17 & 0.77 \\
    \hspace{0.7em} + \textbf{Consistency Loss} & \textbf{6.70} & \textbf{0.48} & \textbf{0.65} & \textbf{1.16} & \textbf{0.76} \\
    SVDFormer $^\dagger$\cite{zhu2023svdformercomplementingpointcloud}& 6.54 & 0.48 & 0.70 & 1.30 & 0.83 \\
    \hspace{0.7em} + \textbf{Consistency Loss} & \textbf{6.52} & \textbf{0.47} & \textbf{0.68} & \textbf{1.21} & \textbf{0.79} \\
%    AdaPoinTr \cite{yu2023adapointr} & 0.49 & 0.69 & 1.24 & 0.81 & 6.53\\
    AdaPoinTr $^\dagger$ \cite{yu2023adapointr} & 6.53 & 0.51 & 0.69 & 1.28 & 0.83 \\
    \hspace{0.7em} + \textbf{Consistency Loss} & \textbf{6.51} & \textbf{0.47} & \textbf{0.68} & \textbf{1.24} & \textbf{0.79} \\
    \bottomrule
    \end{tabular}
    \label{tab:result_pcn_shapenet5}
\end{table*}

\begin{table*}[htb!]
    \centering
    \caption{Quantitative results on the ShapeNet34 benchmark. We report the L2-norm Chamfer Distance (CD$_{l2}$). S, M, and H represent the simple, moderate, and hard setups, where the proportions of missing points are 25\%, 50\%, and 75\%, respectively. $\Delta$ is the gap between the mean CDs of the 21 unseen categories and the 34 seen categories.\\}
    \begin{tabular}{lccccccccc}
    \toprule
     & \multicolumn{4}{c}{34 seen categories} & \multicolumn{4}{c}{21 unseen categories} & $\Delta$\\
     \cmidrule(lr){2-5}
     \cmidrule(lr){6-9}
     & S & M & H & Avg. & S & M & H & Avg. &\\
    & \multicolumn{9}{c}{CD$_{l2} \times 10^3$ $\downarrow$}\\
    \cmidrule(lr){1-10}
    %\cmidrule(lr){2-4}
    FoldingNet& 1.86 & 1.81 & 3.38 & 2.35 & 2.76 & 2.74 & 5.36 & 3.62\\
    PCN& 0.84 & 1.26 & 2.37 & 1.49 & 1.41 & 2.28 & 4.63 & 2.77 & 1.28\\
    \hspace{0.3em} + \textbf{Consistency Loss}  & \textbf{0.57} & \textbf{0.96} & \textbf{1.76} & \textbf{1.09} & \textbf{1.07} & \textbf{1.84} & \textbf{3.70} & \textbf{2.20} & \textbf{1.11}\\
    TopNet& 1.77 & 1.61 & 3.54 & 2.31 & 2.62 & 2.43 & 5.44 & 3.50\\
    GRNet& 1.26 & 1.39 & 2.57 & 1.74 & 1.85 & 2.25 & 4.87 & 2.99\\
    SnowflakeNet & 0.60 & 0.86 & 1.50 & 0.99 & 0.88 & 1.46 & 2.92 & 1.75\\
    PoinTr& 0.76 & 1.05 & 1.88 & 1.23 & 1.04 & 1.67 & 3.44 & 2.05 \\
    AXFormNet& 0.76 & 1.14 & 2.11 & 1.33 & 1.30 & 2.06 & 4.36 & 2.57 & 1.24\\
    \hspace{0.3em} + \textbf{Consistency Loss}  & \textbf{0.48} & \textbf{0.84} & \textbf{1.57}  & \textbf{0.96} & \textbf{0.92} & \textbf{1.67} & \textbf{3.50} & \textbf{2.03} & \textbf{1.07}\\
    SeedFormer& 0.48 & 0.70 & 1.30 & 0.83 & 0.61 & 1.07 & 2.35 & 1.34\\
    PointAttn& 0.51 & 0.70 & 1.23 & 0.81 & 0.76 & 1.15 & 2.23 & 1.38 & 0.57\\
    \hspace{0.3em} + \textbf{Consistency Loss}  & \textbf{0.47} & \textbf{0.66} & \textbf{1.15} & \textbf{0.76} & \textbf{0.61} & \textbf{1.00} & \textbf{2.23} & \textbf{1.28} & \textbf{0.57}\\
    SVDFormer& 0.46 & 0.65 & 1.13 & 0.75 & 0.61 & 1.05 & 2.19 & 1.28 & 0.53\\
    \hspace{0.3em} + \textbf{Consistency Loss}  & \textbf{0.46} & \textbf{0.64} & \textbf{1.12} & \textbf{0.75} & \textbf{0.61} & \textbf{0.98} & \textbf{2.21} & \textbf{1.27} & \textbf{0.52}\\
    AdaPoinTr& 0.51 & 0.68 & 1.09 & 0.76 & 0.63 & 1.06 & 2.23 & 1.30 & 0.54\\
    \hspace{0.3em} + \textbf{Consistency Loss}  & \textbf{0.46} & \textbf{0.62} & \textbf{1.09} & \textbf{0.72} & \textbf{0.63} & \textbf{1.03} & \textbf{2.25} & \textbf{1.30} & \textbf{0.58}\\
    \bottomrule
    \end{tabular}
    
    \label{tab:result_shapenet34}
\end{table*}

\begin{table}[ht]
  \centering
  \caption{Completion Performance of various Point Completion Network with proposed Consistency Loss on MVP Dataset. Model with symbol $^\dagger$ indicates that the models are trained from scratch based on source codes from \cite{zhu2023svdformercomplementingpointcloud} and \cite{yu2023adapointr} \\}
  \begin{tabular}{lcc}
      \toprule
      Point Completion Model &  CD$_{l2} $ $\times 10^{-3} \downarrow$ & F1-Score$@1\% \uparrow$ \\ %\mathcal{L}^{\text{c-tg}} (\beta)$ & CD$_{l2} \cdot 10^3$\\
      \cmidrule{1-3}
      TopNet \cite{tchapmi2019topnet} & 10.11 & 0.308 \\
      PCN \cite{yuan2018pcn} & 9.77 & 0.321 \\
      
      ECG \cite{Pan2020ECGEP} & 7.25 & 0.434 \\
      CRN \cite{Wang_2020_CVPR_CRN_Cascaded_Refinement_Network_for_Point_Completion} & 6.64 & 0.476 \\
      
      PoinTr $^\dagger$ \cite{yu2021pointr} & 6.15 & 0.456 \\
      VRCNet \cite{pan2021variationalrelationalpointcompletion_mvp_point_completion_prediction_dataset} & 5.96 & 0.496 \\
      SVDFormer $^\dagger$ \cite{zhu2023svdformercomplementingpointcloud} & 5.92 & 0.502 \\
      \hspace{0.3em} + \textbf{Consistency Loss} & \textbf{5.73} & \textbf{0.511} \\
      AdaPoinTr $^\dagger$ \cite{yu2023adapointr} & 4.71 & 0.545 \\
      \hspace{0.3em} + \textbf{Consistency Loss} & \textbf{ 4.65} & \textbf{0.553}\\
 
      \bottomrule
  \end{tabular}
  \label{tab:performance_of_point_completion_network_on_MVP_Dataset}
\end{table}

\begin{figure*}[ht]
    \centering
    \includegraphics[width=0.90\textwidth]{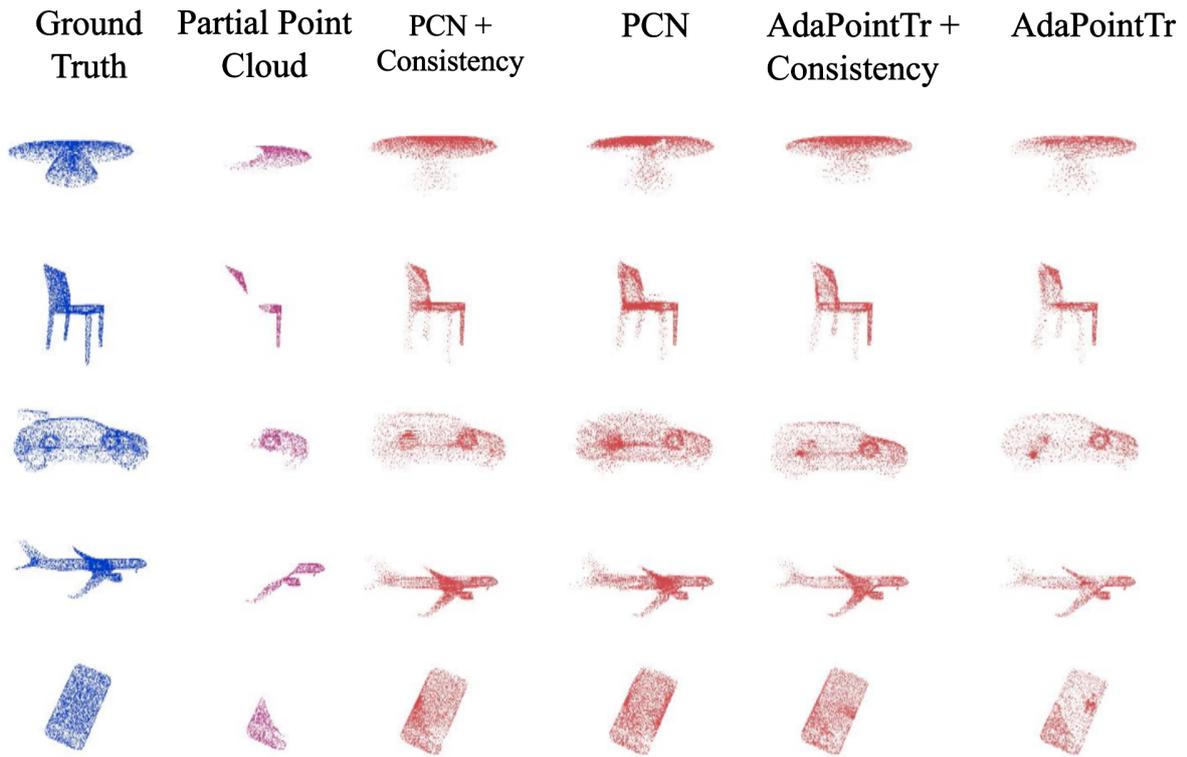}
    \caption{Completion results on the Shapenet55 dataset (test split).}
    \label{fig:shapenet55}
\end{figure*}
\begin{figure*}[!h]
    \centering
    \includegraphics[width=0.90\textwidth]{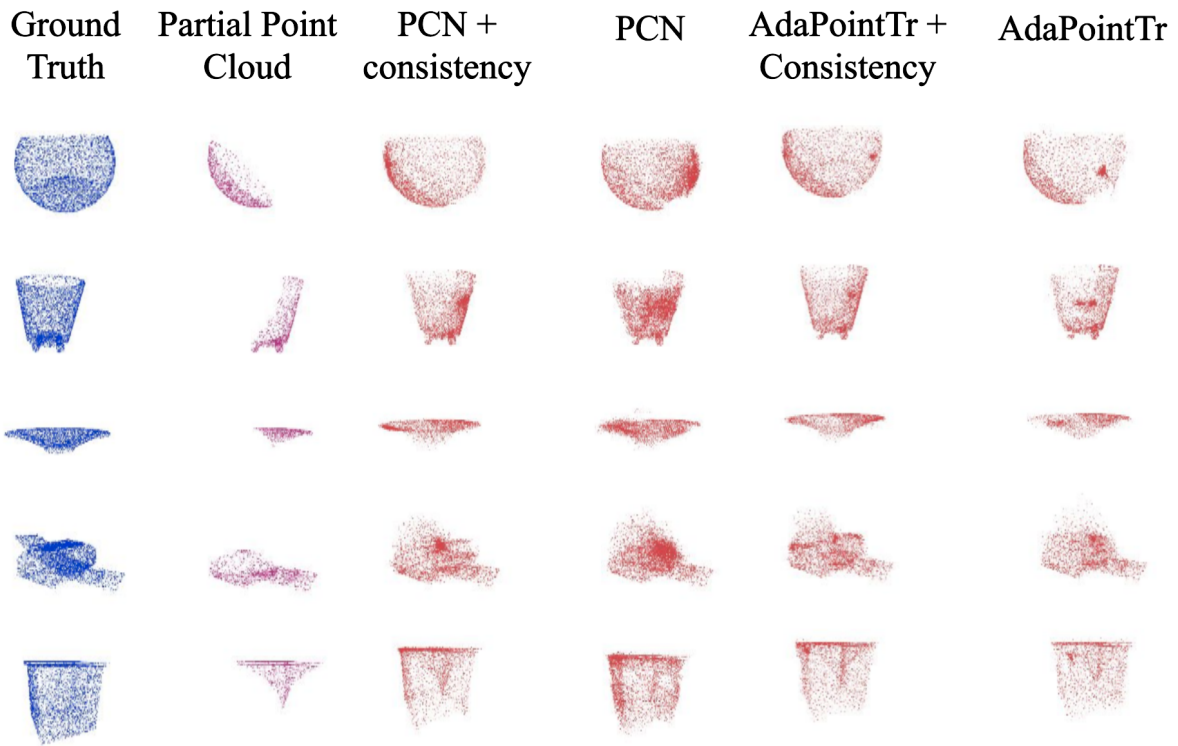}
    \caption{Completion results on the Shapenet34 dataset (test split - \textit{unseen}).}
    \label{fig:shapenet21}
\end{figure*}

\begin{figure}[ht] 
    \centering 
    \includegraphics[width=0.45\textwidth]{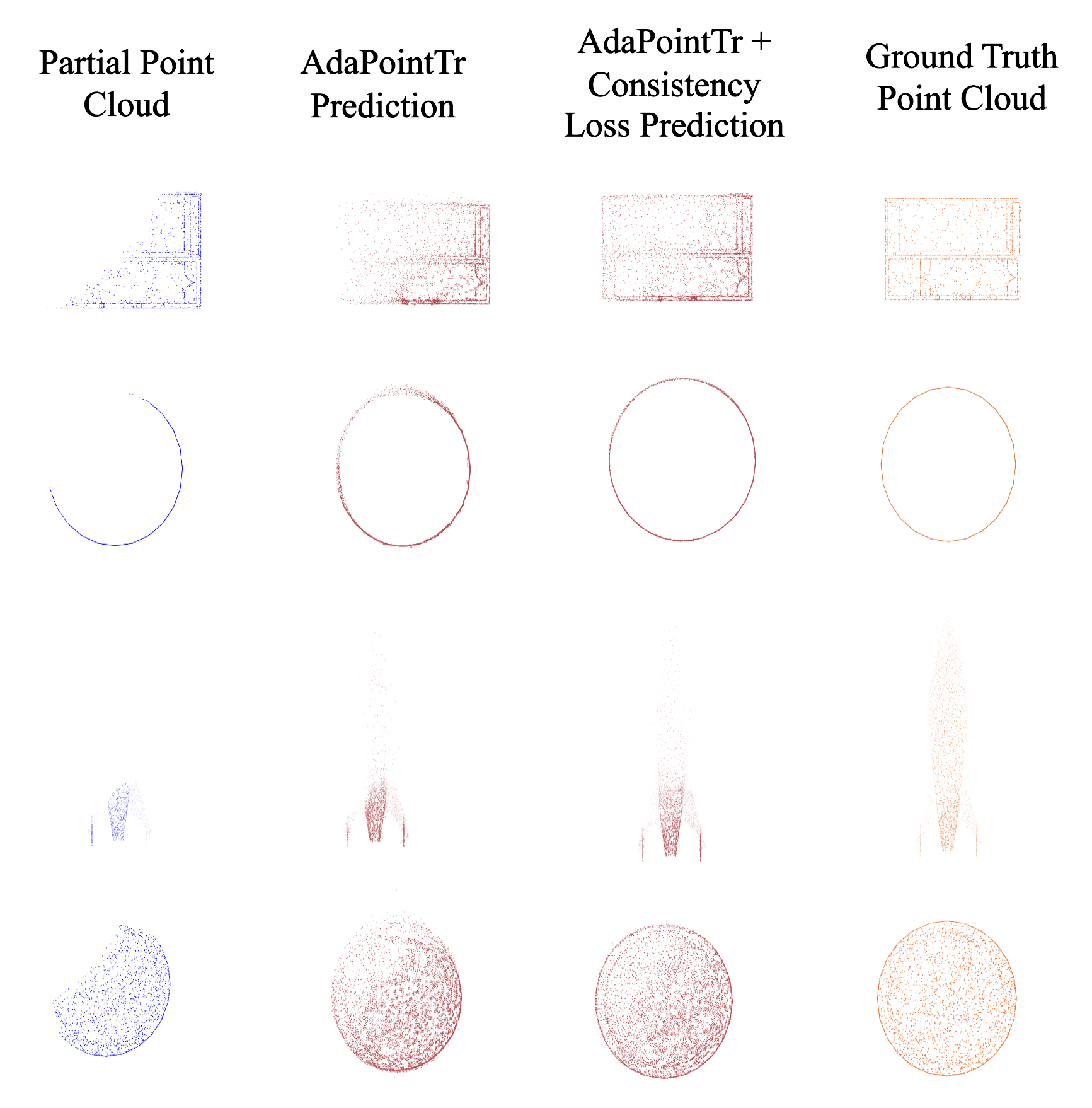} 
    \caption{Comparison of performance for point completion networks AdaPointTr trained with Consistency Loss and without Consistency Loss on challenging dataset MVP Point Completion dataset test split.} 
    \label{fig:comparison_performace_of_point_completion_network_AdapointTr_with} 
\end{figure}

Following \cite{yu2021pointr}, we report the CD$_{l2}$ metric on three difficulty levels for Shapenet55 and the CD$_l1$ metric for PCN in Table \ref{tab:result_pcn_shapenet5}.
From the results, we can draw the following conclusions.

\textbf{The consistency loss improves the completion performance of the three PCCNs across the board.} 
As shown in Table \ref{tab:result_pcn_shapenet5}, the consistency loss can, to some extent, improve the completion performance of the PCCNs in datasets with less diversity such as PCN. However, the consistency loss significantly improves the completion performance of PCN, AxFormNet, and AdaPoinTr on Shapenet55 that consists of objects with diverse geometrical shapes.
Specifically, the completion performance is improved by 27\%, 25\%, and 4.8\% for PCN, AxFormNet, and AdaPoinTr, respectively.
Similar improvements can also be seen on Shapenet34 (Table \ref{tab:result_shapenet34}), in which the mean CD of all three PCCNs trained with the consistency loss are lower or equal to the mean CD of their original counterparts. While in newest point completion dataset MVP dataset we can see that training two state-of-the-art of point completion network AdaPointTr and SVDFormer using consistency loss can increase the performance of point completion network significantly \ref{tab:performance_of_point_completion_network_on_MVP_Dataset} that are $0.19$ CD metric decrease for SVD former a $0.06$ CD metrics decrease for AdaPointTr. This is because MVP dataset contains hight-quality 26 partial point clouds for every point clouds ground truth hence make point completion network trained using proposed consistency loss can learn semantic and geometric information of partial point clouds from same ground truth.

These results demonstrate the effectiveness of the consistency loss for improving the completion performance of existing PCCNs, especially when we are interested in completing a collection of point clouds with diverse geometrical shapes and there are many high quality partial point clouds in same point clouds ground truth.

\textbf{The consistency loss enables fast and accurate point cloud completion.}
Point cloud completion is often used as an auxilliary task, therefore, the completion process should be fast to avoid unnecessary overhead to the overall process.
However, recent PCCNs such as PoinTr \cite{yu2021pointr} and SeedFormer \cite{zhou2022seedformer} achieve improved completion performance at the expense of inference latency due to the complex design of the network.

On the other hand, the proposed consistency loss enables simpler networks to be as accurate as more complex networks, thus improving the completion performance without sacrificing inference latency.
Specifically on the Shapenet55 dataset, PCN with consistency loss achieves a mean CD of $1.07 \cdot 10^{-3}$, which is better than the mean CD of PoinTr ($1.09 \cdot 10^{-3}$).
Another example is the AxFormNet with consistency loss that achieves a mean CD of $0.91 \cdot 10^{-3}$, which is better than the mean CD of SeedFormer ($0.92 \cdot 10^{-3}$).
Considering that, when evaluated on a single RTX 3080Ti GPU, the inference latency of PCN (1.9 ms) and AxFormNet (5.3 ms) are significantly lower than PoinTr (11.8 ms) and SeedFormer (38.3 ms), the consistency loss is a promising training strategy that can enable fast and accurate point cloud completion.

% \textbf{The consistency loss could improve the generalization capability of PCCNs to previously-unseen objects.} 
\textbf{The consistency loss could improve the generalization capability of PCCNs to previously-unseen objects.}
It is desirable for a PCCN to produce accurate completed point clouds even for objects from categories that are not concerned during training.
To quantify the generalization capability of a PCCN, we can consider the gap between the evaluation results on Shapenet34-\textit{seen} split and Shapenet34-\textit{unseen} split, which we refer to as $\Delta$ in Table \ref{tab:result_shapenet34}
% To quantify the generalization capability of a PCCN, we can consider the gap between the evaluation results on Shapenet34-seen split and Shapenet34-unseen split, which we refer to as $\Delta$ in Table \ref{tab:result_shapenet34}
From the table we can see that incorporating the consistency loss results in a significant improvements in the gaps for PCN and AxFormNet, while the gap for AdaPoinTr stays relatively similar. We can also see training point completion network using proposed consistency loss on dataset with many high- quality partial point clouds can increase performance of point completion network to detect unseen points on unseen category \ref{tab:performance_of_point_completion_network_on_MVP_Dataset}.
These results indicate that the consistency loss can act as an additional regularizer for point cloud completion.

\subsubsection{Qualitative Results on Shapenet55 and Shapenet34}
We visualize the completion results of PCN and AdaPointTr on point clouds from the Shapenet55-test and the Shapenet34-\textit{unseen} splits in Figure \ref{fig:shapenet55} and Figure \ref{fig:shapenet21}, respectively.
For each object, we use 25\% of the points in the point cloud as inputs, which is equivalent to the \textit{hard} setup in \cite{yu2021pointr}.
As shown in the figures, networks that are trained with the consistency loss (AdaPointTr+con and PCN+con) predict completed point clouds with equal or better quality compared to the networks that are trained without the consistency loss.
For example, on row 1 in Figure \ref{fig:shapenet55}, AdaPointTr+con can predict the surface of a table with more consistent point density with respect to the ground truth compared to point completion network AdaPointTr trained without proposed consistency loss.
And PCN+con can predict the complete surface of a table, whereas the surface of a table predicted by PCN contains a missing part on the left side.
%While we can see state-of-the-art point completion network AdaPointTr trained on challenging MVP dataset with proposed consistency loss can predict unseen point cloud much accurate than state-of-the-art point completion network AdaPointTr trained without proposed consistency loss. We can see in the result of prediction AdaPointTr trained with proposed consistency loss in \ref{fig:comparison_performace_of_point_completion_network_AdapointTr_with} can predict unseen point cloud much accurate that is can predict unseen point cloud even in not dense point cloud area while point completion network trained without consistency loss couldnt predict unseen point cloud in area with not dense partial point cloud well. This is because challenging MVP dataset contains high-quality 26 multi-view point cloud from same ground truth point cloud hence state-of-the-art point completion network AdaPointTr trained with consistency loss can learn underlying geometric pattern of partial point clouds from same ground truth point clouds.
We observe that the state-of-the-art point completion network AdaPointTr, when trained on the challenging MVP dataset with the proposed consistency loss, predicts unseen point clouds more accurately than AdaPointTr trained without the proposed consistency loss. As shown in the prediction results in \ref{fig
}, AdaPointTr trained with the proposed consistency loss can accurately predict unseen point clouds, even in regions with low point density. In contrast, the point completion network trained without the consistency loss struggles to predict unseen point clouds accurately in such sparse areas. This improvement is due to the challenging MVP dataset, which contains high-quality 26 multi-view point clouds derived from the same ground truth. As a result, the state-of-the-art AdaPointTr network trained with consistency loss can better learn the underlying geometric patterns of partial point clouds from the same ground truth point clouds.

\subsection{Effect of the Number of Points for Consistency Loss on Point Cloud Completion Performance}

\begin{figure}[ht] \centering \includegraphics[width=0.45\textwidth]{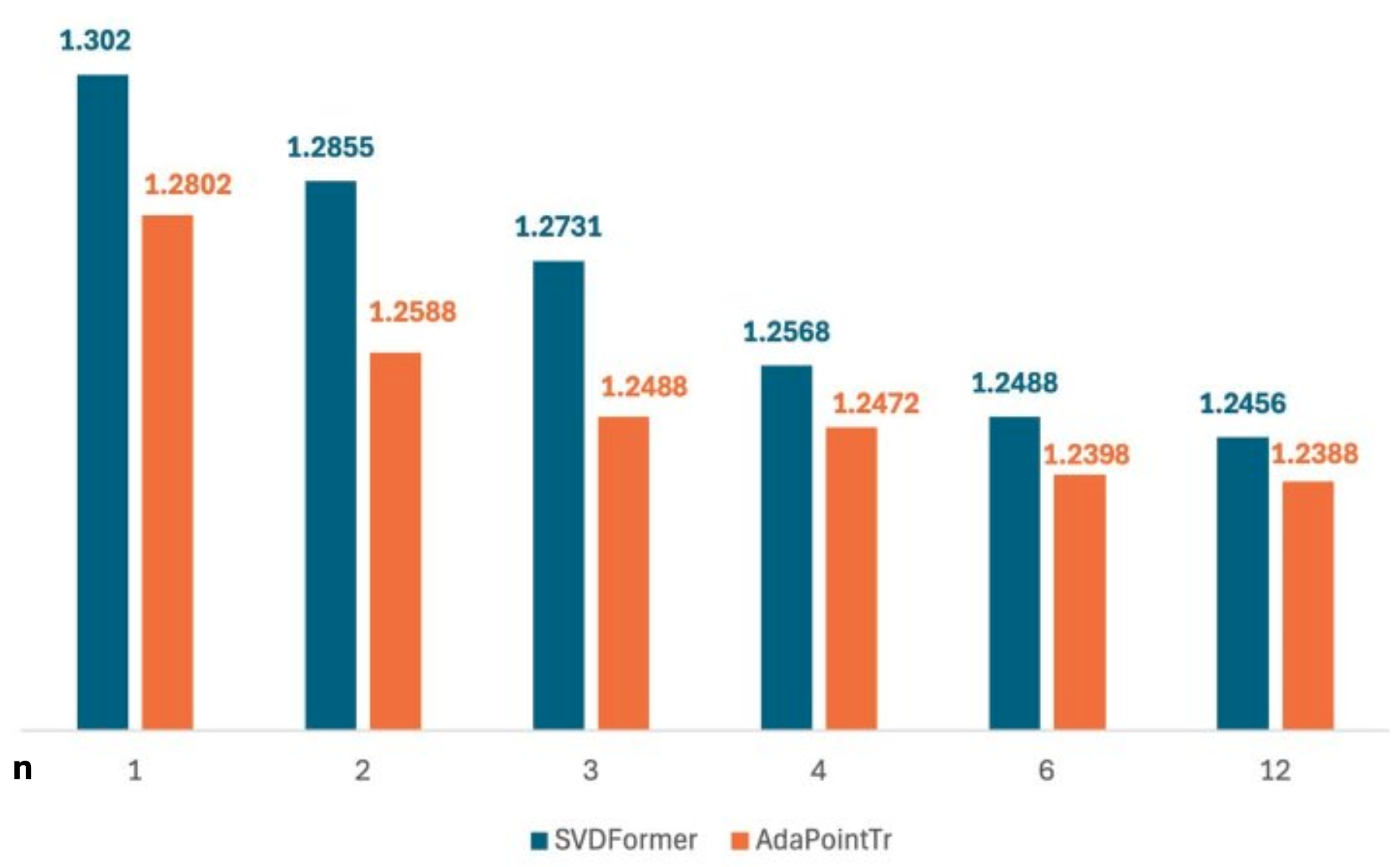} \caption{Comparison of performance for point completion networks SVDFormer and AdaPointTr trained with consistency loss on different numbers of points for consistency loss ($n$) on the ShapeNet-55 dataset.} \label{fig
} \end{figure}

We conducted experiments to evaluate the effect of the number of points used for consistency loss ($n$) on the performance of point cloud completion. Using the ShapeNet-55 dataset with 25% partial point clouds, we investigated how varying the number of points for consistency loss impacts completion performance. For these experiments, we used a batch size of 12, adjusting it based on the number of point clouds for consistency loss: batch size 6 for $n=2$, batch size 4 for $n=3$, and so on, ensuring that both SVDFormer and AdaPointTr trained on 12 point clouds per batch.

As shown in Figure \ref{fig
}, the performance of both SVDFormer and AdaPointTr point completion networks improves as the number of point clouds for consistency loss increases. Specifically, SVDFormer's Chamfer Distance metric improved from 1.302 to 1.2731, and AdaPointTr's improved from 1.2802 to 1.2588. The performance of the point completion networks increased significantly when the number of point clouds for consistency loss was $n = 2, 3,$ and $4$. However, when the number of point clouds for consistency loss was increased to $n = 6$ and $12$, the performance improvement was not significant. This indicates that the optimal number of point clouds for consistency loss, which maximizes performance while minimizing computational cost, is $n=3$.

\subsection{Effect of the Number of Points for Consistency Loss on Point Cloud Completion Training Time}

\begin{figure}[ht] \centering \includegraphics[width=0.45\textwidth]{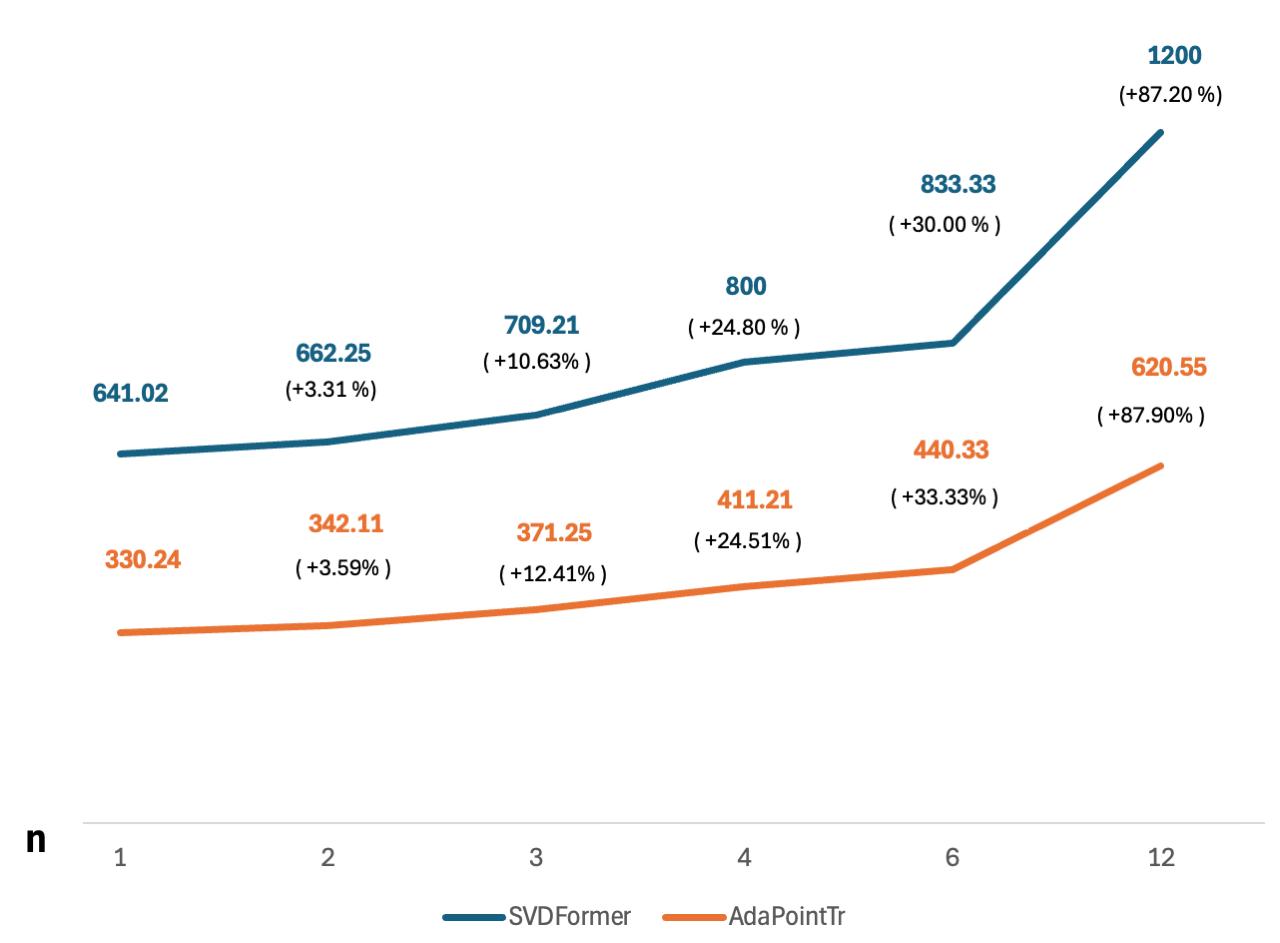} \caption{Comparison of training time for point completion networks trained using the proposed consistency loss with different numbers of points for consistency loss ($n$) on the ShapeNet-55 dataset.} \label{fig
} \end{figure}

We also examined the effect of varying the number of point clouds for consistency loss on the training time of point completion networks. In this experiment, we trained state-of-the-art point completion networks, SVDFormer and AdaPointTr, using a batch size of 12 with different numbers of point clouds for consistency loss on the ShapeNet-55 dataset. We used a batch size of 6 for $n=2$, a batch size of 4 for $n=3$, and so forth, ensuring that both SVDFormer and AdaPointTr were trained on 12 different point clouds per batch.

Figure \ref{fig
} shows that training the point completion networks with the proposed consistency loss led to a modest increase in training time. For instance, the training time of SVDFormer increased from 641.02 ms to 709.21 ms per batch (an increase of approximately 10.63%), and for AdaPointTr, the time increased from 330.24 ms to 371.25 ms (an increase of around 12.41%) when $n=3$. However, when the number of point clouds for consistency loss was increased to $n=12$, the training time for SVDFormer increased significantly from 641.02 ms to 1200 ms (an 87.20% increase), and for AdaPointTr, it increased from 330.24 ms to 620.55 ms (an 87.90% increase).

This significant increase in training time is due to the additional computation required for the proposed self-guided consistency loss, which involves calculating the correlation of point completion predictions with every prediction from the same ground truth data. As the number of point clouds for consistency loss increases, these calculations grow quadratically, leading to longer training times. From these experiments on the effects of the number of points for consistency loss on both point completion performance and training time, we conclude that the best balance between performance improvement and training efficiency is achieved when the point cloud completion networks are trained with $n=3$ point clouds for consistency loss.

% \subsection{Effect of the Noise in Point Completion Network using Consistency Loss}
% Then we trained state-of-the-art Point Completion Network PCN \cite{yuan2018pcn}, SVDFormer \cite{zhu2023svdformercomplementingpointcloud} and AdaPointTr \cite{yu2023adapointr} to predict missing point clouds using incomplete point clouds with generated random noise on benchmark point completion dataset PCN dataset \cite{yuan2018pcn}. The result of point completion network trained on incomplete point cloud with generated random noise can be seen in table \ref{tab:performance_of_point_completion_network_on_random_noise_PCN_dataset}. From performance of state-of-the-art point completion network PCN , SVDFormer and AdaPointTr on incomplete point cloud with random noise on point cloud completion dataset PCN Dataset we can see that performance of state-of-the-art PCN , SVDFormer and AdaPointTr on incomplete point cloud using generated noise have been decreasing significantly since the incomplete point cloud with generated noise is hard to predict than incomplete point cloud without generated noise. We can also see that state-of-the-art Point Completion network trained on incomplete point clouds using generated noise with proposed consistency loss can predict missing point clouds more accurate. This can be explained because proposed consistency loss can make state-of-the-art point completion network extract feature of incomplete point clouds with generated noise from same ground truth more accurate than state-of-the-art point completion network trained without proposed consistency loss.

\subsection{Effect of Noise in the Point Completion Network Using Consistency Loss}

We trained state-of-the-art models, including the Point Completion Network (PCN) \cite{yuan2018pcn}, SVDFormer \cite{zhu2023svdformercomplementingpointcloud}, and AdaPointTr \cite{yu2023adapointr}, to predict missing points in incomplete point clouds augmented with random noise. This training was conducted on the benchmark Point Completion dataset, the PCN dataset \cite{yuan2018pcn}. The performance of these models trained with noisy, incomplete point clouds is shown in Table \ref{tab:performance_of_point_completion_network_on_random_noise_PCN_dataset}.

From the results, we observe that the performance of the state-of-the-art models PCN, SVDFormer, and AdaPointTr decreases significantly when trained on noisy, incomplete point clouds compared to training on incomplete point clouds without noise. This decrease is expected, as the added noise makes it more challenging to accurately predict missing points.

However, when we introduce the proposed consistency loss during training, the models perform better at predicting missing points in noisy point clouds. This improvement can be attributed to the consistency loss, which helps the models extract features from noisy, incomplete point clouds with greater accuracy, resulting in predictions that are more consistent with the ground truth.

\begin{table}[ht]
  \centering
  \caption{Performance of state-of-the-art Point Completion Network trained on point completion dataset PCN dataset with generated noise in incomplete point cloud. Model with symbol $^\dagger$ indicates that the models are trained from scratch based on source codes from \cite{yuan2018pcn},\cite{zhu2023svdformercomplementingpointcloud} and \cite{yu2023adapointr} \\}
  \begin{tabular}{lcc}
      \toprule
      Point Completion Model $^\dagger$ &  CD$_{l2} $ $\times 10^{-3} \downarrow$ \\ %\mathcal{L}^{\text{c-tg}} (\beta)$ & CD$_{l2} \cdot 10^3$\\
      \cmidrule{1-2}
      PCN \cite{yuan2018pcn} & 11.12 \\
      \hspace{0.3em} + \textbf{Consistency Loss} & \textbf{10.01} \\
      SVDFormer $^\dagger$ \cite{zhu2023svdformercomplementingpointcloud} & 7.21  \\
      \hspace{0.3em} + \textbf{Consistency Loss} & \textbf{6.56} \\
      AdaPoinTr $^\dagger$ \cite{yu2023adapointr} & 7.10  \\
      \hspace{0.3em} + \textbf{Consistency Loss} & \textbf{ 6.32} \\
 
      \bottomrule
  \end{tabular}
  \label{tab:performance_of_point_completion_network_on_random_noise_PCN_dataset}
\end{table}

\subsection{Additional Results}
In the following subsection we show additional results from experiments with AxFormNet to further investigate the effects of the consistency loss.
We limit the scope of the experiments to the hardest setup of ShapeNet55 during training and evaluation.

\textbf{Scaling Factors for $\mathcal{L}^{\text{c-sg}}$ and $\mathcal{L}^{\text{c-tg}}$.}
We also investigate the effect of scaling factors $\alpha$ and $\beta$ in Equation \ref{eq:loss_function} as shown in Table \ref{tab:ablation_scaling}.
As a baseline, we use the AxFormNet network trained to predict the missing point clouds as in Table \ref{tab:axform_missing_prediction}, this is equivalent to $\alpha = \beta = 0$. 
First, we investigate the individual effect of each component in the consistency loss.
From the table we can see that both $\mathcal{L}^{\text{c-tg}}$ ($\beta = 1$) and $\mathcal{L}^{\text{c-sg}}$ ($\alpha = 1$)  improve the completion accuracy, with $\mathcal{L}^{\text{c-tg}}$ bringing more benefits compared with $\mathcal{L}^{\text{c-sg}}$.
However, when both are used with the same scaling factors (i.e., $\alpha = \beta = 1$), the completion accuracy is worse than when only $\mathcal{L}^{\text{c-tg}}$ is used.
From experimental results, we see that setting $\alpha = 0.1$ and $\beta = 1$ yield the best completion accuracy.

\begin{table}[ht]
  \centering
  \caption{Completion performance of various AxFormNet on ShapeNet55-hard where 75\% of the original points are missing.\\}
  \begin{tabular}{ccc}
      \toprule
      $\mathcal{L}^{\text{c-sg}} (\alpha)$ & $\mathcal{L}^{\text{c-tg}} (\beta)$ & CD$_{l2} \cdot 10^3$\\
      \cmidrule{1-3}
      0 & 0 & 1.62 \\
      0 & 1 & 1.51 \\
      1 & 0 & 1.60 \\
      1 & 1 & 1.54 \\
      0.1 & 1 & 1.48 \\
      \bottomrule
  \end{tabular}
  \label{tab:ablation_scaling}
\end{table}

\textbf{Number of Training Samples.}
To implement the consistency loss, we sample \textit{n} instances of incomplete point clouds per object to be fed to the PCCN.
This means that the network has access to \textit{n} times more number of samples during training.
A natural question would raise: is the completion accuracy gain simply a result of more training data?
To answer this question, we train the original AxFormNet on Shapenet55 with extra budgets, that is, increasing the number of training epochs to 1200, a threefold increase.
We find that the original AxFormNet trained with extra budgets achieves a $\text{CD}_{l2} \times 10^3$ score of 1.60, which is worse than AxFormNet trained with the consistency loss ($\text{CD}_{l2} \times 10^3 = 1.48$).
This result indicates that the completion performance gains in networks trained with the consistency loss are not simply the results of more training data.

%In the experiment We can see that point completion network trained using proposed consistency loss can predict missing point clouds better on incomplete point clouds using generated noise. Then we can further experiment of performance of point completion network trained using proposed consistency loss for real-world point clouds dataset such as SUN RGB-D Dataset \cite{SUN_RGB_D_Dataset_RGB_D_Benchmark_Dataset} and KITTI Dataset \cite{KITTIDataset_Are_we_ready_for_autonomous_driving}. In the experiment We also see that training point completion network with high number of point from same ground truth point cloud can increase performance of point cloud completion hence We can also create parallel training to make training point completion network using proposed consistency loss much fast and more accurate. We can also try to experiment point completion network trained using proposed consistency loss to detect and classify 3D objects.

In our experiments, we observed that the point completion network trained with the proposed consistency loss better predicts missing points in incomplete point clouds with generated noise. We can further extend these experiments to evaluate the performance of this approach on real-world point cloud datasets, such as the SUN RGB-D Dataset \cite{SUN_RGB_D_Dataset_RGB_D_Benchmark_Dataset}, KITTI Dataset \cite{KITTIDataset_Are_we_ready_for_autonomous_driving}, KRadar Dataset \cite{paek2022k_k_radar_dataset}, NuScene \cite{caesar2020nuscenes} and Waymo Open Dataset \cite{KITTIDataset_Are_we_ready_for_autonomous_driving}. Additionally, we found that training the network with a high number of points from the same ground truth point cloud enhances point cloud completion performance. This suggests that parallel training could further improve both the speed and accuracy of the point completion network when using the proposed consistency loss. Future work could also explore training the network to detect and classify 3D objects using real-world 3D point clouds using the consistency loss approach.
\section{Conclusion}
\label{sec:sec5}

We have proposed the completion consistency loss, a novel loss function for point cloud completion.
The completion consistency loss has been designed to reduce the adverse effects of contradictory supervision signals by considering multiple incomplete views of a single object in one forward-backward pass.
We have demonstrated that the completion consistency loss can improve the completion performance and generalization capability of existing point cloud completion networks without any modification to the design of the networks.
Moreover, simple and fast point cloud completion networks that have been trained with the proposed loss function can achieve completion performance similar to more complex and slower networks.
Therefore, the completion consistency loss can pave the way for accurate, fast, and robust point cloud completion networks.

\section*{CRediT authorship contribution statement}
\textbf{Kevin Tirta Wijaya:} Conceptualization, Data curation, Formal analysis, Methodology, Software, Validation, Visualization, Writing – original draft. \textbf{Christofel Rio Goenawan:} Conceptualization, Formal Analysis, Investigation, Software, Validation, Visualization, Writing – review \& editing. \textbf{Seung-Hyun Kong:} Conceptualization, Funding acquisition, Project administration, Resources, Supervision.

\section*{Declaration of competing interest}

The authors declare that they have no known competing financial interests or personal relationships that could have appeared to influence the work reported in this paper.

\section*{Data availability}
Data and experiment code and result will be made available on request.

\section*{Acknowledgments}

This work was supported by the National Research Foundation of Korea (NRF) grant funded by the Korea government (MSIT) (No. 2021R1A2C3008370).

\appendix
\section{Generating Toy Datasets}

\label{sec:generating_toy_datasets}
The toy datasets that are used in Subsection \ref{sec:pitfall} are generated by following Algorithm \ref{alg:dataset_generator}.
CD is the chamfer distance function defined in Equation $\ref{eq:cd_metric}$.

\newpage

\begin{algorithm}
\caption{Generating Toy Datasets}
\label{alg:dataset_generator}
    \begin{algorithmic}
        \State \textbf{Input}: Full dataset ${\mathbb{D}}$
        \State Initialize ${\mathbb{D}}^A$ as an empty tensor, $k_1 \gets 100$, $k_2 \gets 5$, $n \gets 5000$
        \While{$\text{len}({\mathbb{D}}^A) \leq n$}
            \State Sample ${\bm{X}} \text{ from } {\mathbb{D}}$           
            \State Initialize ${\mathbb{D}}^{\text{inc}}$, ${\mathbb{D}}^{\text{mis}}$, ${\mathbb{D}}^{\text{inc}}$, ${\mathbb{D}}^{\text{mis}}$ as empty tensors.
            \For{${\bm{Y}} \text{ in } {\mathbb{D}}$}
                \State Append $\text{CD}({\bm{X}}^{\text{inc}}, {\bm{Y}}^{\text{inc}})$ to ${\mathbb{D}}^{\text{inc}}$
            \EndFor
            \State Calculate $k_1$-lowest CD-metric in ${\mathbb{D}}^{\text{inc}}$
            \State Append the $k_1$ corresponding ${\bm{Y}} \in {\mathbb{D}}$ to ${\mathbb{D}}^{\text{inc}}$

            \For{${\bm{Z}} \text{ in } {\mathbb{D}}^{\text{inc}}$}
                \State Append $\text{CD}({\bm{X}}^{\text{mis}}, {\bm{Z}}^{\text{mis}})$ to ${\mathbb{D}}^{\text{mis}}$
            \EndFor
            \State Calculate $k_2$-highest CD-metric in ${\mathbb{D}}^{\text{mis}}$           
            \State Append the $k_2$ corresponding ${\bm{Z}} \in {\mathbb{D}}$ to ${\mathbb{D}}^{mis}$           
            
            \If{${\bm{X}} \notin {\mathbb{D}}^A$}
                \State Append ${\bm{X}} \in {\mathbb{D}}$ to ${\mathbb{D}}^A$
            \EndIf

            \For{${\bm{Z}} \text{ in } {\mathbb{D}}^{\text{mis}}$}
                \If{${\bm{Z}} \notin {\mathbb{D}}^A$}
                    \State Append ${\bm{Z}}$ to ${\mathbb{D}}^A$
                \EndIf
            \EndFor            
        \EndWhile

        \State Return the first $n$ elements in ${\mathbb{D}}^A$
    \end{algorithmic}
\end{algorithm}

\bibliographystyle{elsarticle-num}
\bibliography{main}

%% For numbered reference style

% \end{thebibliography}
\end{document}

%% file: math_command.tex
\usepackage{amsmath,amsfonts,bm}

\def\eqref#1{equation~\ref{#1}}
\def\1{\bm{1}}

\DeclareMathAlphabet{\mathsfit}{\encodingdefault}{\sfdefault}{m}{sl}
\SetMathAlphabet{\mathsfit}{bold}{\encodingdefault}{\sfdefault}{bx}{n}

\newcommand{\E}{\mathbb{E}}

\newcommand{\R}{\mathbb{R}}